\documentclass[runningheads]{llncs}

\usepackage{eccv}

\usepackage{eccvabbrv}

\usepackage{graphicx}
\usepackage{booktabs}

\usepackage[accsupp]{axessibility}  

\usepackage{hyperref}

\usepackage{orcidlink}
\usepackage{colortbl}

\begin{document}

\title{Dual-Path Adversarial Lifting for Domain Shift Correction in Online Test-time Adaptation} 

\titlerunning{DPAL}

\author{Yushun Tang\inst{1,2}\orcidlink{0000-0002-8350-7637} \and
Shuoshuo Chen\inst{1}\orcidlink{0000-0001-5689-5788} \and
Zhihe Lu\inst{2}\orcidlink{0000-0002-6917-8654}\and
\\Xinchao Wang\inst{2\star}\orcidlink{0000-0003-0057-1404}\and
Zhihai He\inst{1,3\star}\orcidlink{0000-0002-2647-8286}}

\authorrunning{Y. Tang et al.}

\institute{Southern University of Science and Technology, Shenzhen, China \\
 \and
National University of Singapore, Singapore
 \and
Pengcheng Laboratory, Shenzhen, China\\
\email{\{tangys2022, chenss2021\}@mail.sustech.edu.cn}\\
\email{zhihelu.academic@gmail.com,}
\email{xinchao@nus.edu.sg,}
\email{hezh@sustech.edu.cn}
}

\maketitle

\renewcommand{\thefootnote}{\fnsymbol{footnote}}
\footnotetext[1]{Corresponding authors.} 
\renewcommand{\thefootnote}{\arabic{footnote}}

\begin{abstract}
Transformer-based methods have achieved remarkable success in various machine learning tasks. How to design efficient test-time adaptation methods for transformer models becomes an important research task. In this work, motivated by the dual-subband wavelet lifting scheme developed in multi-scale signal processing which is able to efficiently separate the input signals into principal components and noise components, we introduce a dual-path token lifting for domain shift correction in test time adaptation. Specifically, we introduce an extra token, referred to as \textit{domain shift token}, at each layer of the transformer network. We then perform dual-path lifting with interleaved token prediction and update between the path of domain shift tokens and the path of class tokens at all network layers. The prediction and update networks are learned in an adversarial manner. Specifically, the task of the prediction network is to learn the residual noise of domain shift which should be largely invariant across all classes and all samples in the target domain. In other words, the predicted domain shift noise should be indistinguishable between all sample classes. On the other hand, the task of the update network is to update the class tokens by removing the domain shift from the input image samples so that input samples become more discriminative between different classes in the feature space. To effectively learn the prediction and update networks with two adversarial tasks, both theoretically and practically, we demonstrate that it is necessary to use smooth optimization for the update network but non-smooth optimization for the prediction network.  Experimental results on the benchmark datasets demonstrate that our proposed method significantly improves the online fully test-time domain adaptation performance. Code is available at \url{https://github.com/yushuntang/DPAL}.
\end{abstract}

\section{Introduction}
\label{sec:intro}

\begin{figure}[!ht]
    \centering
    \includegraphics[width = 0.95\textwidth]{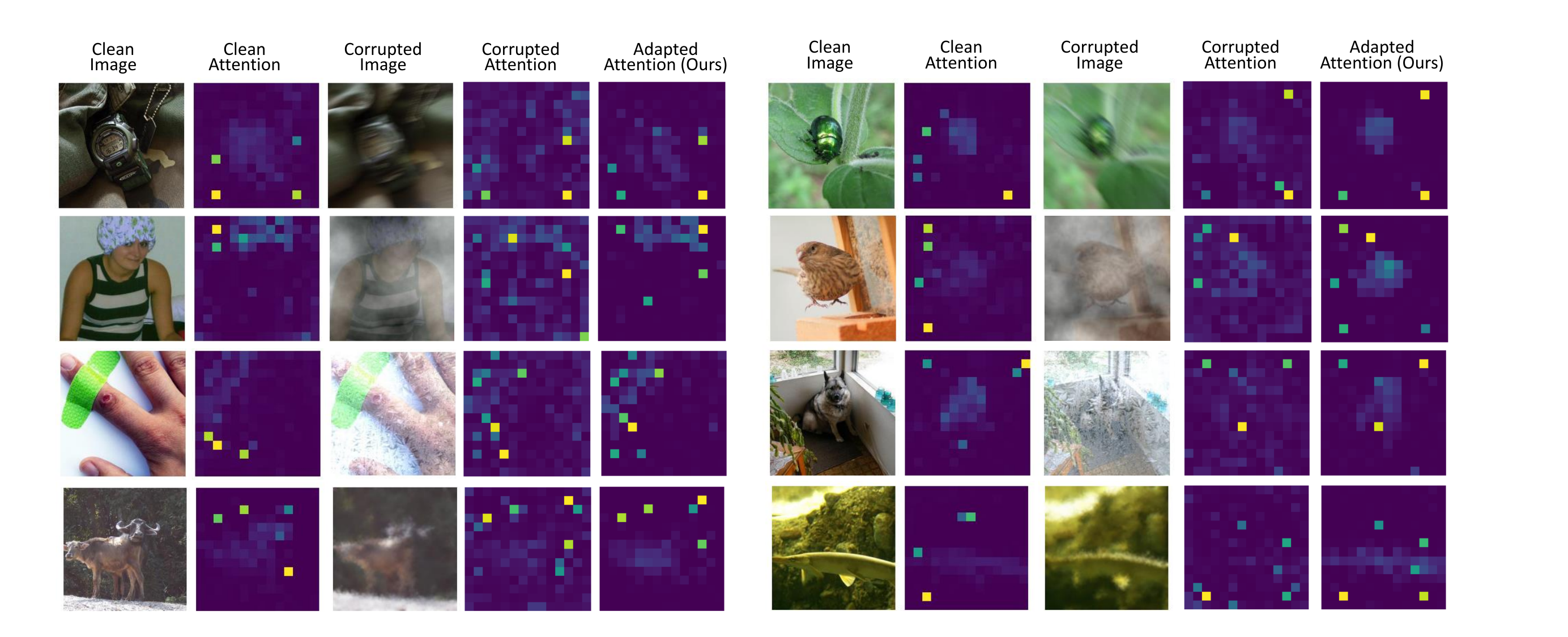}
    \caption{Representative examples of attention weights of the last layer in ViT-B/16 for different corruptions. The class token without adaptation focuses on almost the whole image due to the domain shift. The class token after our adaptation focuses on the object that is close to the clean image attention. There are also some peaky outlier values as explained in \cite{darcet2023vision}.}
    \label{fig: introduction}
\end{figure}

Domain adaptation emerges as a pivotal challenge within computer vision applications, particularly due to the performance degradation observed when models, initially trained on a source domain, are subsequently applied to a target domain that exhibits large domain shifts \cite{mirza2021robustness,lu2020stochastic,zhang2024hiker}.
This challenge has catalyzed significant interest in adapting pre-trained models from the source domain to unlabeled data in the target domain. This research task is referred to as \textit{source-free unsupervised domain adaptation} \cite{liang2020we,Wang_2022_CVPR,li2020model,limttn,tang2023cross,lu2023uncertainty} or \textit{test-time adaptation} (TTA) \cite{wang2020tent,mirza2022norm,niu2022efficient,wang2022continual,tang2023neuro,yuan2023robust,tang2024learning,chen2024learning,kan2023self,yu2023distribution,tang2024domain}.
In this work, we focus on fully test-time adaptation where the model is updated online during inference and the model only has one-time access to the unlabeled data.

Recently, transformer-based methods have achieved remarkable success in various machine learning tasks due to their powerful self-attention capabilities. 
Existing domain adaptation methods have been largely focusing on ResNet-like network models \cite{Sunttt,liu2021ttt++,wang2020tent,mirza2022norm}. Test-time adaptation of transformer models has not been adequately studied and remains a challenging research task.
In our studies, we observe that domain shifts have a significant impact on the self-attention maps in the transformer network and the major challenge in adapting transformer models is to remove the noise from the self-attention map corrupted by the domain shifts. 
Fig. \ref{fig: introduction} shows 
an example where the self-attention map has been severely corrupted in the target domain while our proposed method is able to largely remove domain shift and restore the original self-attention map. Now, the research question becomes: how can we separate the domain shift or corruptions from the self-attention map?

Drawing inspiration from the wavelet lifting scheme, an effective method for noise separation in signal processing \cite{liu2011adaptive,han2013intelligent,hattay2013geometric,zhang2020fast}, we introduce a novel method termed \textit{Dual-Path Adversarial Lifting}. The lifting scheme involves a three-step process: signal splitting, noise prediction, and updating the new signal.
Specifically, we introduce an extra new token positioned ahead of the class token, referred to as \textit{domain shift token}, at each layer of the transformer network. 
We then perform dual-path lifting with interleaved token prediction and update between the path of domain shift tokens and the path of class tokens at all network layers in an online test-time adaptation setting. 

The task of the prediction network is to learn to predict the shift or corruption between the source domain and the target domain. Note that this domain shift is universal across all samples from different classes in the target domain. 
In other words, the predicted domain shift noise should be indistinguishable between all sample classes at each layer of the ViT.
The task of the update network aims to correct the class token so that the class tokens are more distinctive between classes.
Given these two adversarial tasks, 
motivated by the domain adversarial learning method in \cite{rangwani2022closer}, we employ smooth optimization for the update network and non-smooth optimization for the prediction network.
This dual-path optimization strategy enhances the learning of domain shift information. Consequently, our method facilitates the layer-by-layer removal of domain shift information during online testing, allowing the model to maintain its performance across diverse domains.
Our extensive experimental results demonstrate that the proposed Dual-Path Adversarial Lifting transformer significantly improves the online test-time domain adaptation performance and outperforms existing state-of-the-art methods by large margins.

\section{Related Work and Major Contributions}
\label{sec:related_work}
This work is related to test-time adaptation, source-free unsupervised domain adaptation, and prompt learning. 

\subsection{Test-time Adaptation}
Test-time adaptation (TTA) aims to adapt a pre-trained source model to unlabeled data with domain shift during inference. 
Sun et al.~\cite{Sunttt} propose the first test-time training (TTT) method which updates the feature extractor parameters using a self-supervised loss. Based on TTT, the TTT-MAE method \cite{gandelsman2022test} incorporates a transformer backbone and replaces the rotation prediction with masked auto-encoders \cite{he2022masked}. The TTT++ method \cite{liu2021ttt++} improves this approach with a feature alignment strategy based on online moment matching. Note that all these TTT methods require specialized training in the source domain. 
In contrast, fully online test-time adaptation methods fine-tune given pre-trained source models during inference without specialized training in the source domain. For example, the TENT method \cite{wang2020tent} proposes fully test-time adaptation by fine-tuning Batch Normalization (BN) layers. 
The DDA method \cite{gao2022back} projects the input data from the target domain into the source domain based on a diffusion model during testing. 
The NHL method \cite{tang2023neuro} refines early-layer feature representations by unsupervised Hebbian learning.
The SAR \cite{niutowards} method proposed to eliminate noisy samples with large entropy and perform flattening of model weights towards a minimum, enhancing the robustness of the model. The DELTA method \cite{zhao2023delta} uses moving average statistics to perform online adaptation of the normalized features. The SoTTA method \cite{gong2024sotta} filters out noisy samples by high-confidence uniform-class sampling.
In this work, we perform the same task setting following these methods for fully online test-time adaptation.

\subsection{Source-free Unsupervised Domain Adaptation}
Source-free unsupervised domain adaptation (source-free UDA) aims to adapt the model trained on the source domain to the unlabeled target domains without leveraging the source data \cite{liang2020we,sun2022prior,Yang_2021_ICCV,li2020model,lee2013pseudo,huang2021model}. 
Pseudo-labeling \cite{lee2013pseudo} methods assign a class label for each unlabeled target sample and uses the label for the supervised learning objective. 
The SHOT method~\cite{liang2020we} computed pseudo labels through the nearest centroid classifier and optimized the model with information maximization criteria.
The KUDA method \cite{sun2022prior} utilized the prior knowledge of label distribution to refine model-generated pseudo labels.
The SFDA-DE method~\cite{ding2022source} aligned domains by estimating source class-conditioned feature distribution. 
The HCL method~\cite{huang2021model} introduced both instance-level and category-level historical contrastive learning for adaptation. The DIPE method~\cite{Wang_2022_CVPR} explored the domain-invariant parameters of the network model. These source-free methods are offline, requiring access to the complete test dataset. It also costs a number of epochs for model adaptation. 
In contrast, our fully online test time adaptation adapts the given source model on the fly during testing which only accesses the test samples once.

\subsection{Prompt Learning}
Prompt learning has been successfully applied to vision-language models  \cite{zhou2022coop, zhou2022conditional, shu2022test, yu2023task, li2024graphadapter,zhang2024cross,zhang2024concept,zhang2023bdc}. CoOp \cite{zhou2022coop} fine-tunes the prompt of the text encoder in CLIP \cite{radford2021clip}. CoCoOp \cite{zhou2022conditional} conditions the learned prompt on the model's input data to address out-of-distribution issues. TPT \cite{shu2022test} optimizes the prompt of CLIP's text encoder during test time, enhancing generalization performance by minimizing entropy with confidence selection. 
In addition to text prompt learning, techniques have been developed for learning visual prompts in computer vision tasks. In \cite{gan2023decorate},  domain-specific and domain-agnostic prompts are learned and attached to target input images on a per-pixel basis. The BlackVIP method \cite{oh2023blackvip} learns individual prompts for each image through a neural network, without requiring prior knowledge about the pre-trained model.
Visual Prompt Tuning methods \cite{jia2022visual,gao2022visual,sun2023vpa} introduce task-specific learnable parameters into the input sequence of each ViT encoder layer while keeping the pre-trained transformer encoder backbone frozen during downstream training. The VPA method \cite{sun2023vpa} incorporates both prependitive and additive learnable prompts. This approach has found applications in transfer learning for image synthesis \cite{sohn2022visualgan}. Note that these visual prompt tuning methods introduce a large number of new tokens. It substantially increases the computational complexity of the self-attention mechanism, as the computational complexity of self-attention is quadratic to the number of input tokens. Recent literature \cite{darcet2023vision} reveals that the newly introduced tokens besides the class token in ViT is able to attend to different parts of the feature map. In this paper, we incorporate one single additional domain shift token to learn the domain shift in the different target domains. 
Table \ref{tab:related} summarizes the difference between the related works and our method.

\begin{table}[ht]
    \centering
    \caption{Comparison between our method and the related works.}
{
    \begin{tabular}{l|c|c|c}
    \toprule
          Method & Number of prompts & Prependitive prompts & Additive prompts \\
    \midrule
          VPT \cite{jia2022visual} & 10 - 100 & $\checkmark$ & $\times$ \\
          VPA \cite{sun2023vpa} & 50 & $\checkmark$  & $\checkmark$\\
          Ours & 1 & $\checkmark$ &$\times$\\
    \bottomrule
    \end{tabular}  
    }
    \label{tab:related}
\end{table}

\subsection{Major Contributions}
The {major contributions} of this work can be summarized as follows: (1) We introduce a dual-path token lifting for progressive correction of domain shift in test time adaptation. 
Specifically, we introduce an extra token, referred to as \textit{domain shift token}, at each layer of the transformer network. We then perform dual-path lifting with interleaved token prediction and updating between the path of domain shift tokens and the path of class tokens at all network layers. 
(2) We learn the domain shift prediction task and the classification task by dual-level optimization in an adversarial manner. Specifically, we employ smooth optimization for the update network but non-smooth optimization for the prediction network to enhance the learning of domain shift information. 
(3) Our experimental results demonstrate that our proposed Dual-Path Adversarial Lifting method is able to significantly improve the adaptation performance of transformer models, outperforming the state-of-the-art method in fully online test-time domain adaptation across multiple popular benchmark datasets.

\section{Method}
\label{sec:methods}

In this section, we present our method of Dual-Path Adversarial Lifting for domain shift correction in online test-time adaptation.

\subsection{Method Overview}
\label{sec: overview}
Suppose we are given a source trained model $\mathcal{M} = f_{\theta_s}(y|X_s)$ with parameters $\theta_s$, successfully trained on source data $\{X_s\}$ with corresponding labels $\{Y_s\}$. During fully test-time adaptation, we are provided with target data $\{X_t\}$ along with unknown labels $\{Y_t\}$.
In this scenario, we receive a sequence of input sample batches $\{\mathbf{B}_1, \mathbf{B}_2, ..., \mathbf{B}_T\}$ from the target data $\{X_t\}$. It should be noted that during each adaptation step $t$, the network model can only rely on the $t$-th batch of test samples, denoted as $\mathbf{B}_t$.

\begin{figure}[!ht]
    \centering
    \includegraphics[width = 0.99\textwidth]{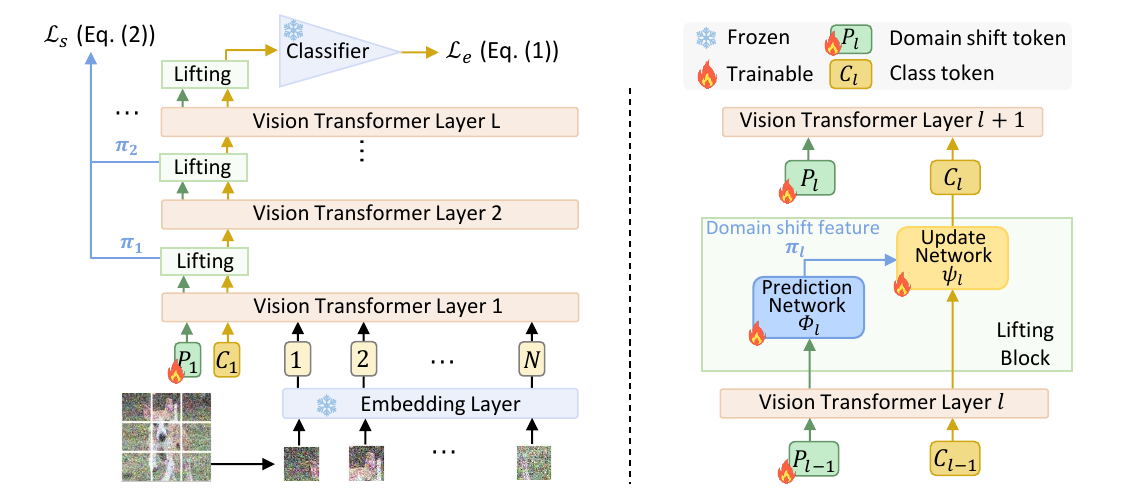}
    \caption{An overview of the proposed Dual-Path Adversarial Lifting method. During inference in the target domain, the domain shift token $P_l$, the prediction network $\Phi_{l}$, and the update network $\Psi_{l}$ are updated given each mini-batch testing samples. The dual-path lifting transformer (\textbf{Left}). The details of the lifting block in each layer (\textbf{Right}).}
    \label{fig: framework}
\end{figure}

There are three steps in the lifting scheme for signal processing: Split, Prediction, and Update. In the splitting step, a signal is divided into two separate signals. The prediction aims to eliminate noise after the initial splitting. Subsequently, the update step corrects the signal by removing this noise. After these steps, input signals are separated into principal components and noise components. 
Inspired by the lifting scheme, in this work, we introduce a dual-path token lifting method for domain shift correction in online test-time adaptation.
When adapting a transformer-based model with  $L$ layers and token dimension $d$ to new domains online efficiently, as shown in Fig. \ref{fig: framework}, we introduce only an additional domain shift token $\mathbf{P} = \{P_1, P_2, \cdots, P_L\}$ ahead of the class token $\mathbf{C} = \{C_1, C_2, \cdots, C_L\}$ at each layer of the Vision Transformer (ViT). Subsequently, we introduce a lifting block including lightweight MLP prediction network $\Phi = \{\Phi_1, \Phi_2, \cdots, \Phi_L\}$ and update network $\Psi = \{\Psi_1, \Psi_2, \cdots, \Psi_L\}$.
The prediction network aims to predict the domain shift feature $\pi_l$ at $l$-th layer. With the domain shift feature, we use the update network to compute the final feature by removing the domain shift feature from the original class token feature. 
Assume that the transformer encoder splits the image feature into two outputs in layer $l$: the domain shift token output $f_l^d$ and the class token output $f_l^c$. The Prediction Network $\Phi_l$ performs domain shift prediction, denoted as $\pi_l=\Phi_l(f_l^d)$. Subsequently, the Update Network $\Psi_l$ performs class token update for the next layer, denoted as $C_{l+1}=\Psi_l(f_l^c,\pi_l)$.
During the forward process of the online test-time adaptation, the outputs $y_t$ for current batch samples $\mathbf{B}_t$ are calculated using the model parameters from the last batch, denoted as $\theta_{t-1}$. Simultaneously, the objective loss is calculated and backward for updating the model parameters to $\theta_{t}$ for the subsequent batch samples $\mathbf{B}_{t+1}$. 
The online test-time adaptation is completed by iteratively executing the forward and backward process in the target domain dataset.
In the following sections, we will further explain the proposed method in more details.

\subsection{Dual-path Token Lifting for Domain Shift Correction}

We introduce two loss functions to guide the optimization for the online adaptation process, named reliable entropy loss $\mathcal{L}_e(\theta_t; x)$ and similarity loss $\mathcal{L}_s(\theta_t; x)$.
During testing, for samples $x$ in the current mini-batch $\mathbf{B}_t$, the reliable entropy loss is defined as:
\begin{equation}\label{eq: loss}
     \mathcal{L}_e(\theta_t; x) = \mathbf{I}\cdot \mathbb{E}(\theta_t; x),
\end{equation}
where $\mathbf{I} = \mathbb{I}[\mathbb{E}(\theta_t; x)<E_0]$ is the mask to filter out test samples whose entropy is large than the threshold $E_0$, $\mathbb{E}$ is the entropy function. 

For the ViT model with $L$ layers, the similarity loss is defined as:
\begin{equation}
    \mathcal{L}_{s}(\theta_t; x) = -\frac{1}{L}\sum\limits_{l=1}^{L} \frac{1}{B^2} \sum\limits_{j=1}^{B} \sum\limits_{k=1}^{B} M_{l}^{jk},
\end{equation}
where $B$ is the batch size, $M$ represents the cosine similarity matrix calculated by each pair of domain shift feature $\pi_l$ obtained from the prediction network in the mini-batch, and $\pi_l  \in \mathcal{R}^{B\times d}$.

On top of the entropy loss $\mathcal{L}_e(\theta_t; x)$, the combination of the similarity loss $\mathcal{L}_{s}(\theta_t; x)$ is used to encourage the prediction network to learn the domain-specific information, which leads to the following loss function:
\begin{equation}
    \mathcal{L}(\theta_t; x) =  \mathcal{L}_{e}(\theta_t; x) + \lambda \cdot \mathcal{L}_{s}(\theta_t; x).
\end{equation}
Here, $\lambda = \frac{\sum \mathbf{I}}{B}$ serves as the balance parameter, where $\sum \mathbf{I}$ denotes the summation of the entropy mask over the batch. 

The prediction networks and update networks are learned in an adversarial manner.   
Specifically, the task of the prediction network is to learn the residual noise of domain shift which should be largely invariant across all classes and all samples in the target domain.
In other words, the predicted domain shift noise should be indistinguishable between all sample classes.
On the other hand, the task of the update network is to update the class tokens by removing the domain shift from the input image sample and patches so that input samples become more discriminative between different classes in the feature space. 
Inspired by domain adversarial learning \cite{rangwani2022closer}, to effectively learn the prediction network and update networks with two adversarial tasks, we employ smooth optimization for the prediction network but non-smooth optimization for the update network. 
In the following section, we explain the smooth and non-smooth optimization of the class token update networks and domain shift prediction networks, respectively.

\subsection{Smooth Optimization for Class Token Update Networks}
We follow the SAR method \cite{niutowards} to perform the Sharpness-Aware Minimization (SAM) \cite{foret2021sharpnessaware} for the update networks.
From the perspective of generalization and optimization, SAM not only minimizes individual points within the loss landscape criterion but also consistently reduces the loss in their surrounding neighborhoods. 
Contrary to the conventional approach of exclusively optimizing the model weights with low loss values, SAM seeks to identify smoother minima in the weight space. These minima are characterized by uniformly low loss $\mathcal{L}$ in the neighborhoods $\epsilon$ of model weights $\theta$:
\begin{equation}
    \underset{\theta}{\min} \;\underset{||\epsilon|| \leq \rho}{\max}\; \mathcal{L}(\theta + \epsilon; x), 
\end{equation}
where $\rho \geq 0$ is a hyper-parameter to define the scope of the neighborhoods. 
To address this minimax problem, SAM initially addresses the maximization problem by seeking the maximum perturbation $\epsilon_t$ at training step $t$. This inner maximization problem can be approximated using the first-order Taylor expansion of $\mathcal{L}(\theta + \epsilon; x)$ with respect to $\epsilon \rightarrow 0$ as follows:
\begin{equation}
\begin{split}
            {\epsilon}_t(\theta) &=\underset{||\epsilon|| \leq \rho}{\arg \max} \;  \mathcal{L}(\theta + \epsilon; x)\\
            &= \underset{||\epsilon|| \leq \rho}{\arg \max} \;  \mathcal{L}(\theta; x) + \epsilon^{\mathbf{T}}\nabla_{\theta} \mathcal{L}(\theta; x) + \mathrm{o}(\epsilon) \\
            &\approx \underset{||\epsilon|| \leq \rho}{\arg \max} \; \epsilon^{\mathbf{T}}\nabla_{\theta} \mathcal{L}(\theta; x). 
\end{split}
\end{equation}
The value $\epsilon_t(\theta)$ that solves this approximation is given by the solution to a classical dual norm problem:
\begin{equation}
    \epsilon_t(\theta) = \rho \cdot sign(\nabla_{\theta} \mathcal{L}(\theta; x))\frac{| \nabla_{\theta} \mathcal{L}(\theta; x) |^{q-1}}{\| \nabla_{\theta} \mathcal{L}(\theta; x) \|_{q}^{q/p}},
\end{equation}
where $\frac{1}{p} + \frac{1}{q} = 1$. It is empirically confirmed that the  optimization yields the best performance when $p = 2$, resulting in $\epsilon_t$ formulated as:
\begin{equation}\label{eq: epsilon}
    \epsilon_t(\theta) = \rho \frac{\nabla_{\theta} \mathcal{L}(\theta; x)}{||\nabla_{\theta} \mathcal{L}(\theta; x)||_2}.
\end{equation}
Then the gradient update for the model weights $\theta_t$ is computed as:
\begin{equation}\label{eq: theta}
    \theta_{t+1} = \theta_t - \eta \nabla_{\theta} \mathcal{L}(\theta; x)|_{\theta_t + \epsilon_t}.
\end{equation}
Finally, SAM converges the model weights $\theta$ to a smooth minimum with respect to the loss function $\mathcal{L}$ by iteratively updating equation \ref{eq: epsilon} and equation \ref{eq: theta}.

The SAR method \cite{niutowards} demonstrates that converging to a smooth minimum with respect to entropy loss stabilizes the adaptation process, resulting in improved performance on the target domain. However, recent literature \cite{rangwani2022closer} indicates that converging to smooth minima concerning adversarial loss results in sub-optimal generalization. Motivated by this non-smooth optimization in adversarial learning, in our study, we observe that non-smooth minima with respect to the similarity loss encourage the prediction network to learn better domain shift information, leading to better generalization performance on the target domain. In section \ref{sec: non-smooth}, we give a theoretical analysis for the non-smooth lifting optimization.

\subsection{Non-Smooth Optimization for Domain Shift Prediction Networks}
\label{sec: non-smooth}

We measure the domain shift prediction through the similarity estimation of features by the prediction network $\Phi$ as $S_{\Phi} = \mathbb{E}[\sum{M}]$. A higher degree of similarity indicates a more effective extraction of domain-specific knowledge within the mini-batch samples, which implies a better estimation of the domain shift. The objective is to maximize the similarity estimation $S_{\Phi}$, with $S_{\Phi*}$ representing the optimal similarity. To theoretically analyze the difference in similarity estimation between the smooth version $S_{\Phi''}$ and the non-smooth version $S_{\Phi'}$, we consider $S_{\Phi}$ as an $L$-smooth function \cite{carmon2020lower}, with $\eta$ denoting a small constant.
For one step of gradient updating during testing in a mini-batch, let $\Phi' = \Phi + \eta \mathbf{v}_1$, where $\mathbf{v}_1 = (\nabla S_{\Phi}/||\nabla S_{\Phi}||)$, and $\Phi'' = \Phi + \eta \mathbf{v}_2$, where $\mathbf{v}_2 = (\nabla S_{\Phi}|_{\Phi + {\epsilon}_t(\Phi)}/||\nabla S_{\Phi}|_{\Phi + {\epsilon}_t(\Phi)}||)$.

During online adaptation in the testing stage, where only a single step of the gradient is used for estimating similarity, any differential function can be approximated by the linear approximation for small $\eta$ and gradient direction $\mathbf{v}_1$:
\begin{equation}
    S_{\Phi + \eta \mathbf{v}_1} \approx S_{\Phi} + \eta \mathbf{v}_1 \cdot \nabla {S_{\Phi}}.
\end{equation}
Expressing the dot product between two vectors in terms of norms and the angle $\alpha$ between them:
\begin{equation}
    \mathbf{v}_1 \cdot \nabla S_{\Phi} = || \nabla S_{\Phi} || \; ||\mathbf{v}_1|| \; \cos \alpha, 
\end{equation}
we achieve the steepest value when $\cos \alpha = 1$, which implies $\mathbf{v}_1 =  \frac{\nabla S_{\Phi}|_\Phi}{||\nabla S_{\Phi}|_\Phi||}$. Now, comparing the gradient descent in another direction $\mathbf{v}_2 = \frac{\nabla S_{\Phi}|_{\Phi + \epsilon_t(\Phi)}}{||\nabla S_{\Phi}|_{\Phi + \epsilon_t(\Phi)}||}$ with smooth optimization, the difference in value can be characterized by:
\begin{equation}\label{eq: diff}
     S_{\Phi + \eta \mathbf{v}_1} - S_{\Phi + \eta \mathbf{v}_2}= \eta ||\nabla S_{\Phi}||(1 - \cos \beta), 
\end{equation}
where $\beta$ is the angle between $\mathbf{v}_1: \nabla S_{\Phi}|_\Phi$ and $\mathbf{v}_2: \nabla S_{\Phi}|_{\Phi + \epsilon_t(\Phi)}$. The suboptimality is dependent on the gradient magnitude. Utilizing the following result, we demonstrate that when the optimality gap $S_{\Phi*} - S_{\Phi}$ is large, the difference between the two directions is also large.
For an \textit{L-smooth} function, which is both continuously differentiable and possesses a Lipschitz continuous gradient with a Lipschitz constant denoted by $L$, we derive the following inequality equation based on \cite{rangwani2022closer}:

\begin{equation}
  \frac{1}{2L} \| \nabla S_{\Phi} \|^2 \leq 
 (S_{\Phi*} - S_{\Phi}). 
\end{equation}
By substituting equation \ref{eq: diff} into the above inequality equation, we obtain:

\begin{equation}
\left( \frac{S_{\Phi + \eta \mathbf{v}_2} - S_{\Phi + \eta \mathbf{v}_1}}{\eta(1 - \cos \beta)}\right)^2 \leq 2L (S_{\Phi*} - S_{\Phi})  , 
\end{equation}

\begin{equation}\label{eq: phi'}
(S_{\Phi'} - S_{\Phi''})^2   \leq 2L\eta^2(1 - \cos \beta)^2(S_{\Phi*} - S_{\Phi}).
\end{equation}

\begin{figure}[!t]
    \centering
    \includegraphics[width = 0.9\textwidth]{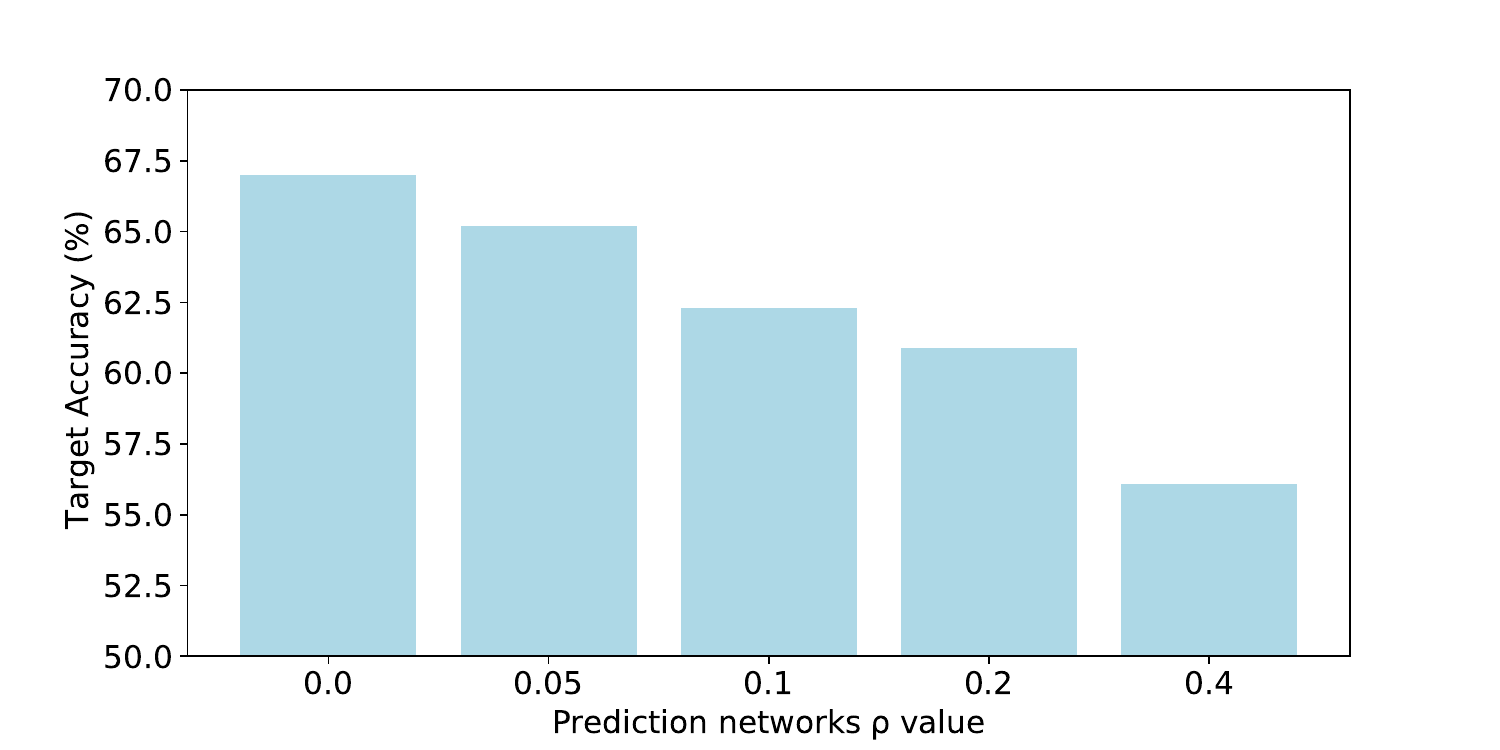}
    \caption{Target domain accuracy for different prediction network smoothness factors. As the smoothness increases ($\rho$), the target accuracy in ImageNet-C decreases, indicating that smoothing similarity loss for the prediction network leads to sub-optimal generalization.}
    \label{fig:predictor_rou}
\end{figure}

Equation \ref{eq: phi'} reveals that the difference between the value obtained by taking a step in the direction of gradient $\mathbf{v}_1$ and taking a step in a different direction $\mathbf{v}_2$ is upper-bounded by $S_{\Phi*} - S_{\Phi}$. Consequently, if we are far from the optimal value, the difference can potentially be large. With only one step of gradient updating during online test-time adaptation, where $S_{\Phi*} - S_{\Phi}$ is potentially large, the smooth version $S_{\Phi''}$ may stand far away from the normal non-smooth gradient-updated similarity estimation, leading to a sub-optimal measure of similarity. In our experiments, as illustrated in Fig. \ref{fig:predictor_rou}, we also observe that increasing the smoothness of the prediction network with respect to the similarity loss (i.e. increasing the scope value $\rho$ of the neighborhoods) leads to lower performance in the target domain. This is because, with increased smoothness in the similarity loss for the prediction network, the prediction network tends to learn more semantic knowledge rather than domain-specific knowledge. This will potentially lead to the loss of crucial semantic information in the subsequent Update Network. 
This loss of semantic information could be potentially large since the upper bound of the difference in equation \ref{eq: phi'} becomes larger, which is detrimental to the classification accuracy. Therefore, a non-smooth optimization for the prediction network gets a better performance both theoretically and practically.
Additional details are provided in the Supplementary Material.

\section{Experimental Results}
\label{sec:exp}

In this section, we conduct experiments on multiple dataset benchmarks to evaluate the performance of our proposed Dual-Path Adversarial Lifting method for online test-time adaptation.

\subsection{Benchmark Datasets and Baselines.} 
In our experiments, we select the widely used \textbf{ImageNet-C} \cite{hendrycks2018benchmarking} benchmark dataset in TTA methods, consisting of $50,000$ instances for each corruption. Additionally, we test in \textbf{ImageNet-R} \cite{hendrycks2021many}, a dataset containing 30,000 images presenting diverse artistic from the ImageNet dataset. The \textbf{ImageNet-A} \cite{hendrycks2021nae} dataset consists of real-world samples that are misclassified by ResNet models.
We also include \textbf{VisDA-2021} \cite{bashkirova2022visda}, a dataset designed to assess models' ability to adapt to novel test distributions.
Additionally, we utilize the \textbf{Office-Home} \cite{venkateswara2017deep} dataset, which has a total of $15,500$ images across four distinct styles.
We compare our proposed Dual-Path Adversarial Lifting method against the following fully test-time adaptation methods: no adaptation which is the source model, T3A, CoTTA, DDA, MEMO, TENT, DePT-G, SAR, and EATA.

\subsection{Implementation Details} 
We use the ViT-B/16 backbone for all experiments. 
The source model weights trained in the ImageNet dataset are obtained from the \textit{timm} repository \cite{rw2019timm}. Specifically, for the Office-Home dataset, we fine-tune the ViT-B/16 model by replacing the original classifier head with a new classifier head as the source model. 
We employ the feature subtracting process and the original normalization module as the update network. The prediction network is a lightweight MLP with one hidden layer and the additional parameter number is about $1.2$ M. The learning rate is set to 1e-2. 
The batch size is set to 64 for all experiments. It should be noted that we use the matched normalization setting for the pre-trained \textit{timm} \cite{rw2019timm} model (mean = [$0.5, 0.5, 0.5$], std = [$0.5, 0.5, 0.5$]), which is different from the code of the original SAR paper \cite{niutowards}. We report the mean and standard deviation values obtained from three runs of testing. All models are trained and tested on a single NVIDIA RTX3090 GPU. 

\begin{table*}[!htbp]
\begin{center}
\caption{Classification Accuracy (\%) for test-time adaptation from ImageNet to each corruption in \textbf{ImageNet-C} at the highest severity (Level 5). All models use the ViT-B/16 backbone, where Source, DDA, MEMO, TENT, and SAR are reproduced by us, while others are cited from \cite{yu2023benchmarking,gao2022visual}. The best result is shown in \textbf{bold}.}
\label{table: normal}
\resizebox{\linewidth}{!}
{
\begin{tabular}{l|ccccccccccccccc|c}
\toprule
Method & gaus & shot & impul & defcs & gls & mtn & zm & snw & frst & fg & brt & cnt & els & px & jpg & Avg.\\	
\midrule
Source & 46.9 & 47.6 & 46.9 & 42.7 & 34.2 & 50.5 & 44.7 & 56.9 & 52.6 & 56.5 & 76.1 & 31.8 & 46.7 & 65.5 & 66.0 & 51.0 \\
T3A \cite{iwasawa2021test} & 16.6 & 11.8 & 16.4 & 29.9 & 24.3 & 34.5 & 28.5 & 15.9 & 27.0 & 49.1 & 56.1 & 44.8 & 33.3 & 45.1 & 49.4 & 32.2\\
CoTTA \cite{wang2022continual} & 40.3 & 31.8 & 39.6 & 35.5 & 33.1 & 46.9 & 37.3 & 2.9 & 46.4 & 59.1 & 71.7 & 55.5 & 46.4 & 59.4 & 59.0 & 44.4\\

DDA~\cite{gao2022back} &   52.5  & 54.0  & 52.1  & 33.8  & 40.6  & 33.3  & 30.2  & 29.7  & 35.0  & 5.0  & 48.6  & 2.7  & 50.0  & 60.0  & 58.8  & 39.1 \\
MEMO~\cite{zhang2022memo} & 58.1 & 59.1 & 58.5 	& 51.6 & 41.2 & 57.1 & 52.4 & 64.1 & 59.0 & 62.7 & \textbf{80.3} & 44.6 & 52.8 & 72.2 & 72.1 & 59.1 \\
AdaContrast \cite{chen2022contrastive} & 54.4 & 55.8 & 55.8 & 52.5 & 42.2 & 58.7 & 54.3 & 64.6 & 60.1 & 66.4 & 76.8 & 53.7 & 61.7 & 71.9 & 69.6 & 59.9 \\
CFA \cite{kojima2022robustvit} & 56.9 & 58.0 & 58.1 & 54.4 & 48.9 & 59.9 & 56.6 & 66.4 & 64.1 & 67.7 & 79.0 & 58.8 & 64.3 & 71.7 & 70.2 & 62.4 \\
TENT~\cite{wang2020tent} & 57.6 & 58.9 & 58.9 & 57.6 & 54.3 & 61.0 & 57.5 & 65.7 & 54.1 & 69.1 & 78.7 & 62.4 & 62.5 & 72.5 & 70.6 & 62.8 \\
DePT-G \cite{gao2022visual} &53.7 & 55.7 & 55.8 & 58.2 & 56.0 & 61.8 & 57.1 & 69.2 & 66.6 & 72.2 & 76.3 & 63.2 & 67.9 & 71.8 & 68.2 & 63.6\\
SAR~\cite{niutowards} & 58.0 & 59.2 & 59.0 & 58.0 & 54.7 & 61.2 & 57.9 & 66.1 & 64.4 & 68.6 & 78.7 & 62.4 & 62.9 & 72.5 & 70.5 & 63.6 \\ 
EATA \cite{niu2022efficient} & 54.8 & 55.3 & 55.6 & 58.0 & 59.1 & 63.4 & 61.5 & 67.7 & 66.2 & \textbf{73.2} & 77.9 & \textbf{68.0} & 68.4 & 73.1 & 70.3 & 64.8 \\
\rowcolor{gray!20}
\textbf{Ours} & \textbf{59.8} & \textbf{61.7} & \textbf{61.0} & \textbf{59.1} & \textbf{60.5} & \textbf{64.9} & \textbf{63.8} & \textbf{70.2} & \textbf{68.9} & 72.6 & 79.7 & 62.6 & \textbf{70.9} & \textbf{75.6} & \textbf{73.1} & \textbf{67.0} \\
\rowcolor{gray!20}
& ${\pm0.1}$ & ${\pm0.1}$ & ${\pm0.0}$ & ${\pm0.3}$ & ${\pm0.2}$ & ${\pm0.1}$ & ${\pm0.1}$ & ${\pm0.2}$ & ${\pm0.3}$ & ${\pm0.3}$ & ${\pm0.1}$ & ${\pm0.4}$ & ${\pm0.1}$ & ${\pm0.0}$ & ${\pm0.0}$ & ${\pm0.1}$ \\ 
\bottomrule
\end{tabular}
}
\end{center}
\end{table*}

\begin{table}[!ht]
    \centering
    \caption{Classification Accuracy (\%)  for test-time adaptation from ImageNet to \textbf{ImageNet-R}, \textbf{ImageNet-A}, and \textbf{VisDA-2021} datasets.}
    \label{tab: imagenet-r}
    {
    \begin{tabular}{l|ccc}
    \toprule
        Method &  ImageNet-R &  ImageNet-A & VisDA-2021\\
    \midrule
        Source &  57.2 &31.1  &  57.7\\
        TENT \cite{wang2020tent}& 61.3 &44.5 &  60.1\\
        SAR \cite{niutowards}& 62.0 &45.3  & 60.1 \\
        \rowcolor{gray!20}
        \textbf{Ours} &   \textbf{64.8} &\textbf{49.9} & \textbf{64.1} \\
        \rowcolor{gray!20}
        &${\pm0.3}$& ${\pm0.5}$ &${\pm0.3}$ \\
    \bottomrule
    \end{tabular}   }
\end{table}

\begin{table*}[!htbp]
\begin{center}
\caption{Classification Accuracy (\%) for test-time adaptation of all transfer tasks in \textbf{Office-Home} dataset.}
\label{table:officehome}
\resizebox{\linewidth}{!}
{
\begin{tabular}{l|cccccccccccc|c}
\toprule
Method & {A$\rightarrow$C} & {A$\rightarrow$P} & {A$\rightarrow$R} & {C$\rightarrow$A} & {C$\rightarrow$P} & {C$\rightarrow$R} & {P$\rightarrow$A} & {P$\rightarrow$C} & {P$\rightarrow$R} & {R$\rightarrow$A} & {R$\rightarrow$C} & {R$\rightarrow$P} & {Avg.} \\
\midrule
Source & 63.4 & 81.9 & 86.3 & 76.2 & 80.6 & 83.8 & 75.0 & 57.9 & 87.2 & 78.7 & 61.0 & 88.0 & 76.7 \\
TENT \cite{wang2020tent} & \textbf{69.1} & 81.8 & 86.5 & 76.5 & 81.9 & 83.2 & \textbf{76.8} & 65.0 & 86.7 & \textbf{81.1} & \textbf{69.7} & \textbf{88.2} & 78.9 \\
SAR~\cite{niutowards} & 67.3 & 80.7 & 85.6 & 77.5 & 79.8 & 84.1 & 74.7 & 60.3 & 87.6 & 78.9 & 63.1 & 87.7 & 77.3 \\
\rowcolor{gray!20}
\textbf{Ours} & 68.6 & \textbf{82.2} & \textbf{86.9 }& \textbf{78.4} & \textbf{83.1} & \textbf{85.3} & \textbf{76.8} & \textbf{65.9} & \textbf{87.7} & 80.7 & 68.7 & 88.4 & \textbf{79.4} \\
\rowcolor{gray!20}
& ${\pm0.3}$ & ${\pm0.2}$ & ${\pm0.0}$ & ${\pm0.2}$ & ${\pm0.2}$ & ${\pm0.1}$ & ${\pm0.3}$ & ${\pm0.1}$ & ${\pm0.1}$ & ${\pm0.1}$ & ${\pm0.1}$ & ${\pm0.1}$ & ${\pm0.0}$ \\
\bottomrule
\end{tabular}
}
\end{center}
\end{table*}

\subsection{Performance Results} 
We first evaluate the performance of our Dual-Path Adversarial Lifting method on the popular TTA benchmark \textbf{ImageNet-C}. We report the reproduced top-1 accuracy for all methods under comparison. 
The results of this experiment are shown in Table \ref{table: normal}. Our method outperforms other baseline methods for almost all 15 corruptions.
On average, our method outperforms the second-best method by 2.2\%. 
We also respectively conduct experiments on the \textbf{ImageNet-R}, \textbf{ImageNet-A}, and \textbf{VisDA-2021} datasets. For ImageNet-R and ImageNet-A, we use the same pre-trained ViT-B/16 backbone and set the output size to 200 following the procedure in \cite{hendrycks2021many, hendrycks2021nae}. From  Table \ref{tab: imagenet-r}, our method outperforms the SAR method by $2.8$\%, $4.6$\%, and $4.0$\% respectively. It demonstrates that the proposed Dual-Path Adversarial Lifting method is effective in different domains.
We extend our experimentation to \textbf{Office-Home}. The results are presented in Table \ref{table:officehome}. The proposed Dual-Path Adversarial Lifting method outperforms the SAR method by 2.1\%. This further underscores the efficacy of our proposed Dual-Path Adversarial Lifting method across different domains with distinct image styles.
Overall, our experimental results consistently demonstrate the effectiveness and robustness of our proposed Dual-Path Adversarial Lifting approach. Moreover, our method outperforms the state-of-the-art TTA methods across multiple evaluation metrics.
More experimental results are provided in the Supplementary Materials.

\subsection{Ablation Studies and  Further Discussions}

\textbf{(a) Contribution of major algorithm components.} We conducted an ablation study, and the results are presented in Table \ref{table:ablation}. When only integrating the domain shift token (DST) without an additional similarity loss, we observed a degradation in performance. This can be attributed to the newly introduced token potentially learning more about semantic information rather than domain-specific information in online adaptation. Consequently, this leads to a reduction in semantic knowledge after the updating process.
Upon incorporating the similarity loss, we observed an improvement in the average accuracy by $1.6$\%. However, the most substantial enhancement, with a boost of $2.4$\%, was achieved when adapting the model with both proposed similarity loss and non-smooth optimization. It demonstrates the significant contribution of all the proposed components in enhancing the model's performance.

\begin{figure}[!ht]
    \begin{minipage}{.47\linewidth}
        \centering
        \captionof{table}{Ablation study in \textbf{ImageNet-C} at the highest severity. The DST represents the domain shift token.}
        \label{table:ablation}
        \begin{tabular}{l|cc}
            \toprule
            Methods  & Avg. \\	
            \midrule
            Baseline  Method (SAR) & 63.6 \\
            \quad + DST   & 62.5\\
            \quad + DST + $\mathcal{L}_s$  & 65.2\\
            \rowcolor{gray!20}
            \textbf{Our  Method}  & \textbf{67.0}\\
            \bottomrule
        \end{tabular}
    \end{minipage}%
    \hspace{0.5cm}
    \begin{minipage}{.47\linewidth}
        \centering
        \captionof{table}{Testing time cost comparison for Gaussian corruption of \textbf{ImageNet-C} with NVIDIA RTX3090 GPU.}
        \label{table:cost}
        \begin{tabular}{l|l}
            \toprule
            Method & Time Cost \\
            \midrule
            SAR & $\sim$ 7 min \\
            MEMO & $\sim$ 14 hours \\
            DDA & $\sim$ 5 days \\
            \textbf{Ours} & $\sim$ 8 min \\
            \bottomrule
        \end{tabular}
    \end{minipage}
\end{figure}

\begin{figure}[!ht]
    \centering
    \includegraphics[width = 0.99\textwidth]{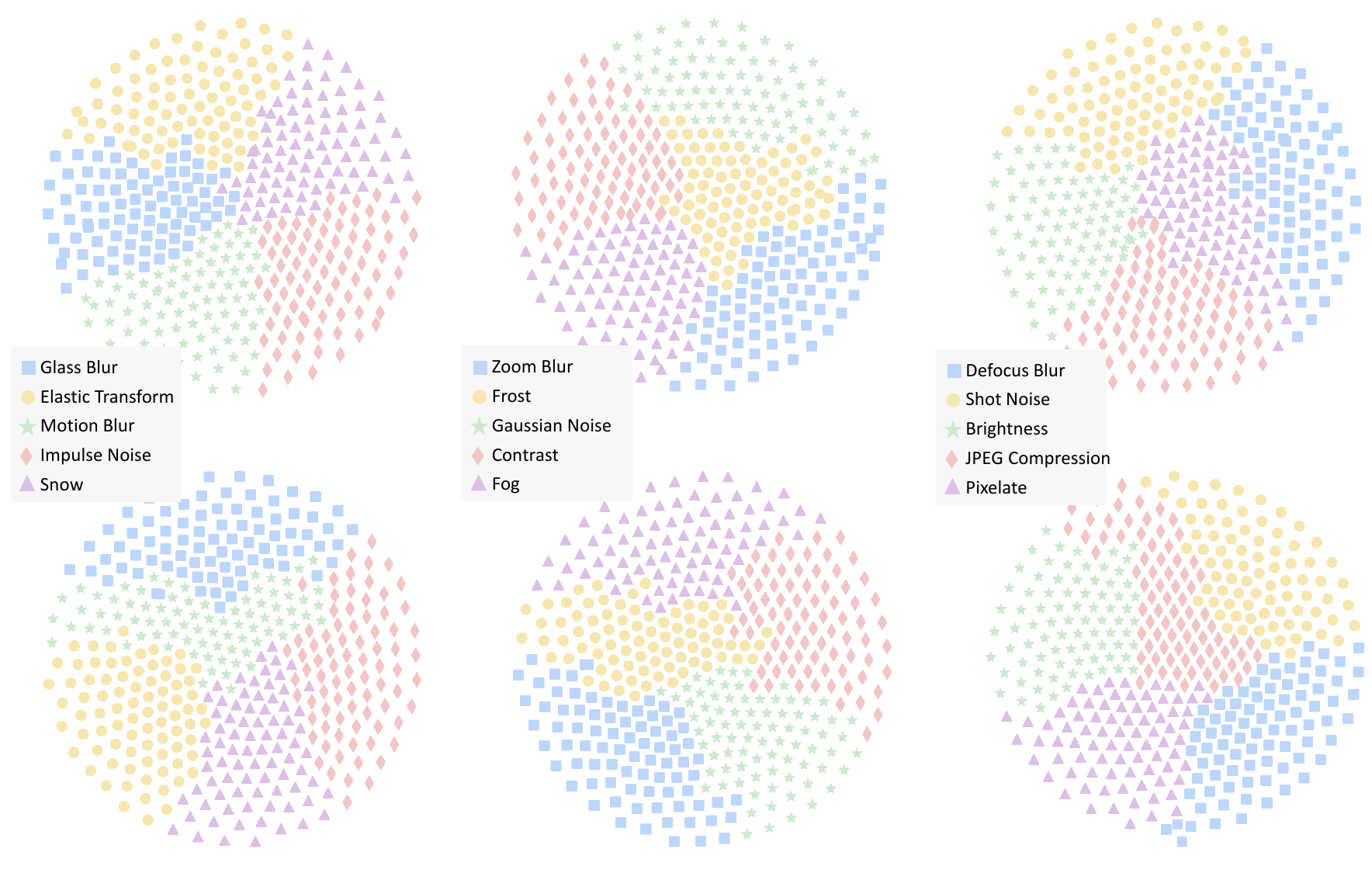}
    \caption{The t-SNE visualization for the domain shift features in the first layer (top) and last layer (bottom) of ViT. Different colors represent different domains. It is able to learn domain-specific knowledge for different domains across layers. }
    \label{fig: TSNE}
\end{figure}

\textbf{(b) Visualization of the prediction features.} In Fig. \ref{fig: TSNE}, we visualize the domain shift features obtained from the prediction networks at the first layer and the last layer in ViT-B/16. Different colors represent different domains. We can see that the features from the prediction networks for the same domain are close together. Note that both the first layer and the last layer exhibit clear and distinguishable features. It effectively demonstrates the capability of the domain shift prediction network to learn domain-specific knowledge across diverse domains at different transformer layers.

\textbf{(c) Running time cost comparison.} In Table \ref{table:cost}, we provide a comprehensive comparison of running time costs for all Gaussian corruption samples of \textbf{ImageNet-C} under the single NVIDIA RTX3090 GPU. Our proposed Dual-Path Adversarial Lifting method introduces a slightly higher computational overhead, approximately 13\% slower than the SAR method, but much faster than MEMO and DDA.
This extra overhead is mainly caused by the learning of the domain shift token and lifting block. The original SAR method only updates the normalization layer parameters. 

\section{Conclusion}
\label{sec:conclusion}
Fully test-time adaptation is a challenging problem in computer vision, particularly in the presence of complex corruptions and shifts in the test data distribution. 
In this work, we introduced a module called Dual-Path Adversarial Lifting to learn the domain-invariant features for domain shift correction.
Specifically, we integrated an extra domain shift token ahead of the class token and introduced a lightweight prediction network to learn the domain-specific information at each layer of the ViT. Once learned the domain-specific information, we updated the final feature by removing the domain-invariant feature from the original class token output feature. The prediction network and update network are learned in an adversarial manner. Furthermore, we employed smooth optimization for the update networks but non-smooth optimization for the prediction networks to enhance the learning of domain information. Then the domain information can be removed online layer by layer, resulting in improved cross-domain performance during the testing stage.
Our experimental results demonstrated that our proposed Dual-Path Adversarial Lifting method is able to significantly improve fully test-time adaptation performance.

\section*{Acknowledgements}
This research is supported by the National Natural Science Foundation of China (No. 62331014) and Project 2021JC02X103,
and the Singapore Ministry of Education Academic Research Fund Tier 1 (WBS: A-8001229-00-00), a project titled ``Towards Robust Single Domain Generalization in Deep Learning''.

%
%
\bibliographystyle{splncs04}
\bibliography{egbib}
\end{document}


\title{Dual-Path Adversarial Lifting for Domain Shift Correction in Online Test-time Adaptation\\Supplementary Materials} 

\titlerunning{DPAL}

\author{Yushun Tang\inst{1,2}\orcidlink{0000-0002-8350-7637} \and
Shuoshuo Chen\inst{1}\orcidlink{1111-2222-3333-4444} \and
Zhihe Lu\inst{2}\orcidlink{0000-0002-6917-8654}\and \\
Xinchao Wang\inst{2\star}\orcidlink{0000-0003-0057-1404}\and
Zhihai He\inst{1,3\star}\orcidlink{0000-0002-2647-8286}}

\authorrunning{Y. Tang et al.}

\institute{Southern University of Science and Technology, Shenzhen, China \\
 \and
National University of Singapore, Singapore
 \and
Pengcheng Laboratory, Shenzhen, China\\
\email{\{tangys2022, chenss2021\}@mail.sustech.edu.cn}\\
\email{zhihelu.academic@gmail.com,}
\email{xinchao@nus.edu.sg,}
\email{hezh@sustech.edu.cn}
}

\maketitle

\appendix
In this Supplementary Materials, we provide more details and experimental results for further understanding of the proposed Dual-Path Adversarial Lifting for Domain Shift Correction in Online Test-time Adaptation.

\section{More Details on the Non-smooth Optimization}
In Section 3.4 of the main paper, we provide the proof of the inequality used in our non-smooth optimization method. Specifically, for an $L$-smooth function $f(w)$, the following inequality holds, where $w^*$ represents the minimum:
\begin{equation}
    f(w) - f(w^*) \geq \frac{1}{2L}  || \nabla f(w) ||^2
\end{equation}

\begin{proof}
The L-smooth function, by definition, satisfies the following inequality:
$$
f(w^*) \leq f(v) \leq f(w) + \nabla f(w) (v - w) + \frac{L}{2}||v - w||^2
$$
Now, to obtain a tight bound on $f(w^*)$, we minimize the upper bound with respect to $v$:
$$
D(v) = f(w) + \nabla f(w) (v - w) + \frac{L}{2}||v - w||^2
$$
Setting $\nabla_{v} D(v) = 0$, we have:
$$
v = w - \frac{1}{L} \nabla f(w) 
$$
Substituting the value of $v$ back into the upper bound yields:
$$
f(w^*) \leq f(w) - \frac{1}{2L} || \nabla f(w) ||^2
$$
Rearranging the above expression, we have:
$$
\frac{1}{2L}  || \nabla f(w) ||^2 \leq f(w) - f(w^*) 
$$

\end{proof}

In the main paper, since we aim for the similarity estimation to be as large as possible, we express it as $f(w) = -S_{\Phi}$. This formulation results in the inequality:
\begin{equation}
    \frac{1}{2L} \| \nabla S_{\Phi} \|^2 \leq 
 (S_{\Phi*} - S_{\Phi}).
\end{equation}

\section{Additional Results on ImageNet-C with ViT-L/16 Backbone}
We extend our experimentation to encompass a larger ViT-L/16 backbone. The batch size is set to 32 due to limited memory. The learning rate is set 
to $0.002$. As shown in Table \ref{tab: vit-l-normal}, our method outperforms the baseline SAR for 4\% in average. This robust performance demonstrates the efficiency of our proposed Dual-Path Adversarial Lifting method across diverse transformer backbones.

\begin{table}[!ht]
\setlength{\abovecaptionskip}{0cm}
    \centering
    \caption{Classification Accuracy (\%)  in \textbf{ImageNet-C} with \textbf{ViT-L/16} at the highest severity (Level 5).}
    \label{tab: vit-l-normal}
    \resizebox{\linewidth}{!}{
\begin{tabular}{l|ccccccccccccccc|c}
\toprule
Method & gaus & shot & impul & defcs & gls & mtn & zm & snw & frst & fg & brt & cnt & els & px & jpg & Avg.\\
    \midrule
        Source & 62.1 & 61.4 & 62.3 & 52.7 & 45.1 & 60.6 & 55.1 & 66.2 & 62.4 & 62.5 & 80.2 & 39.8 & 56.2 & 74.3 & 72.7 & 60.9 \\
        TENT & 65.6 & 68.3 & \textbf{67.6} & \textbf{63.4} & 59.9 & \textbf{66.8} & 60.7 & 69.0 & 68.5 & \textbf{67.4} & 81.0 & 28.9 & 64.7 & \textbf{77.2} & 74.7 & 65.6 \\
        SAR & \textbf{66.0} & 66.6 & 66.2 & 61.3 & 55.1 & 66.1 & 58.3 & 68.4 & 65.7 & 66.3 & 81.0 & 26.8 & 63.7 & 74.5 & 73.6 & 64.0 \\
        \rowcolor{gray!20}
        \textbf{Ours} & 64.5 & \textbf{68.5} & 67.1 & 62.7 & \textbf{60.8} & 66.6 & \textbf{64.1} & \textbf{70.4} & \textbf{69.6} & 66.4 & \textbf{81.2} & \textbf{55.4} & \textbf{69.7} & \textbf{77.2} & \textbf{75.8} & \textbf{68.0}  \\
        \rowcolor{gray!20}
        & ${\pm3.6}$ & ${\pm0.2}$ & ${\pm0.8}$ & ${\pm0.8}$ & ${\pm1.7}$ & ${\pm1.4}$ & ${\pm1.0}$ & ${\pm1.4}$ & ${\pm1.3}$ & ${\pm1.2}$ & ${\pm0.3}$ & ${\pm1.2}$ & ${\pm2.0}$ & ${\pm0.8}$ & ${\pm0.9}$ & ${\pm0.4}$ \\
    \bottomrule
    \end{tabular}   }
\vspace{-0.5cm}
\end{table}

\section{Additional Results on ImageNet-C with Severity Level 3}

We provide additional performance comparison results for corruption severity Level 3 in Table \ref{table: normal_3}. The results are consistent with those in the main paper for severity level 5. We can see that our Dual-Path Adversarial Lifting method consistently outperforms existing methods for all 15 corruption types.

\begin{table}[!htbp]
\setlength{\abovecaptionskip}{0cm}
\begin{center}
\caption{Classification Accuracy (\%) for each corruption in \textbf{ImageNet-C} under \textbf{Normal} at the severity Level 3. The best result is shown in \textbf{bold}.}
\label{table: normal_3}
\resizebox{\linewidth}{!}
{
\begin{tabular}{l|ccccccccccccccc|c}
\toprule
Method & gaus & shot & impul & defcs & gls & mtn & zm & snw & frst & fg & brt & cnt & els & px & jpg & Avg.\\	
\midrule
Source & 72.1 & 71.5 & 71.3 & 62.3 & 51.1 & 69.1 & 59.1 & 69.0 & 60.0 & 70.1 & 80.1 & 74.0 & 75.1 & 77.6 & 75.2 & 69.2 \\
DDA & 62.6  & 63.1  & 62.3  & 50.2  & 54.2  & 50.9  & 43.3  & 41.0  & 41.9  & 14.8  & 60.6  & 26.0  & 62.2  & 62.6  & 62.8  & 50.6 \\
TENT & 74.3 & 73.9 & 73.6 & 70.8 & 66.6 & 73.7 & 66.9 & 73.2 & 68.7 & 76.0 & 81.6 & 78.9 & 78.5 & 79.7 & 77.1 & 74.2 \\
SAR & 74.3 & 73.9 & 73.7 & 70.9 & 66.5 & 73.8 & 66.9 & 73.1 & 68.7 & 75.8 & 81.8 & 78.9 & 78.5 & 79.8 & 77.1 & 74.2 \\ 
\rowcolor{gray!20}
\textbf{Ours} & \textbf{75.1} & \textbf{74.9} & \textbf{74.7} & \textbf{71.9} & \textbf{70.6} & \textbf{75.3} & \textbf{70.7} & \textbf{75.3} & \textbf{72.1} & \textbf{78.0} & \textbf{82.1} & \textbf{79.4} & \textbf{79.9} & \textbf{80.5} & \textbf{79.1} & \textbf{76.0}  \\ 
\rowcolor{gray!20}
& ${\pm0.1}$ & ${\pm0.1}$ & ${\pm0.0}$ & ${\pm0.0}$ & ${\pm0.1}$ & ${\pm0.1}$ & ${\pm0.2}$ & ${\pm0.0}$ & ${\pm0.1}$ & ${\pm0.1}$ & ${\pm0.1}$ & ${\pm0.2}$ & ${\pm0.1}$ & ${\pm0.0}$ & ${\pm0.2}$ & ${\pm0.0}$ \\
\bottomrule
\end{tabular}
}
\vspace{-0.5cm}
\end{center}
\end{table}

\section{Ablation Study on the Number of Domain Shift Tokens}
In this paper, we added one Domain Shift Token (DST) into the transformer. Certainly, we can add more. 
In the following experiment, we study how the number of Domain Shift Tokens affect the overall performance. 
The average accuracy results on ImageNet-C are presented in Table \ref{tab:num_prompt}. We can see that the best performance is achieved when the number of domain shift tokens is set to 1. This is because adding more DST will introduce more parameters which need to be learned by these few samples in the current batch during online test-time adaptation. 

\begin{table}[]
    \centering
    \caption{Classification Accuracy (\%) in \textbf{ImageNet-C} at the highest severity (Level 5) with different number of domain shift tokens.}
    \begin{tabular}{c|cccc}
    \toprule
        Num & 1 & 2 & 4 & 8 \\
    \midrule
        Acc & 67.0 & 66.4 & 62.6 & 66.6 \\
    \bottomrule
    \end{tabular}
    
    \label{tab:num_prompt}
\end{table}



\section{Additional Visualization}
\begin{figure}
    \centering
    \includegraphics[width=\linewidth]{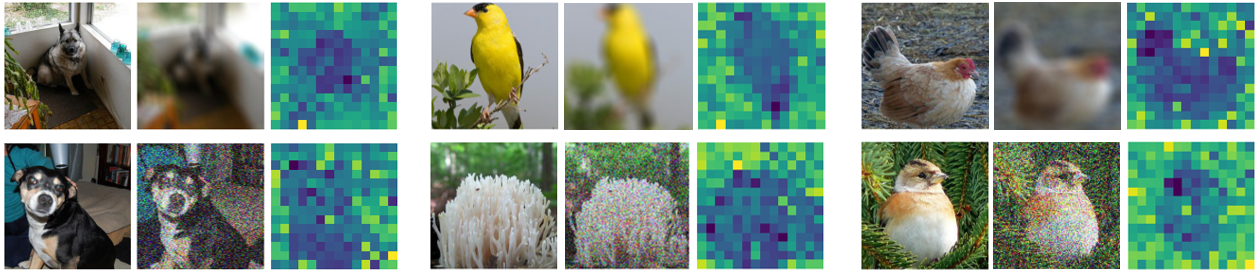}
    \caption{The correlation between the domain shift feature and individual image patches. Clean image (left), Corrupted image (middle), and correlation map (right).}
    \label{fig:correlation}
\end{figure}

We also visualize the correlation between the domain shift feature and individual image patches in Figure \ref{fig:correlation}, where brighter dots represent larger correlation values. We can see that: (1) The domain shift feature has high correlation with almost all patches since the domain shift is across all patches. (2) It has relatively low correlation with the object patches, which is crucial for preserving object semantic information during the subsequent Update Network process.

%
%
\bibliographystyle{splncs04}